\documentclass{article}

\usepackage[preprint]{neurips_2022}
\usepackage[utf8]{inputenc} % allow utf-8 input
\usepackage[T1]{fontenc}    % use 8-bit T1 fonts
\usepackage{hyperref}       % hyperlinks
\usepackage{url}            % simple URL typesetting
\usepackage{booktabs}       % professional-quality tables
\usepackage{amsfonts}       % blackboard math symbols
\usepackage{nicefrac}       % compact symbols for 1/2, etc.
\usepackage{microtype}      % microtypography
\usepackage{xcolor}         % colors
\usepackage{times}
\usepackage{graphicx}
\usepackage{amsmath}
\usepackage{algorithmic}
\usepackage{algorithm}
\usepackage{footnote}
\usepackage{multirow}
\usepackage{makecell}
\usepackage{bm}
\usepackage{amssymb}
\usepackage{enumitem}
\usepackage{epstopdf}
\usepackage{subfigure}
\title{Perturbation Learning Based Anomaly Detection}

\author{%
  Jinyu Cai \\
  Fuzhou University\\
  Shenzhen Research Institute of Big Data\\
  \texttt{jinyucai1995@gmail.com} \\
  \And
  Jicong Fan \\
  The Chinese University of Hong Kong (Shenzhen)\\
  Shenzhen Research Institute of Big Data\\
  \texttt{fanjicong@cuhk.edu.cn} \\
}

\begin{document}

\maketitle

\begin{abstract}
This paper presents a simple yet effective method for anomaly detection. The main idea is to learn small perturbations to perturb normal data and learn a classifier to classify the normal data and the perturbed data into two different classes. The perturbator and classifier are jointly learned using deep neural networks. Importantly, the perturbations should be as small as possible but the classifier is still able to recognize the perturbed data from unperturbed data. Therefore, the perturbed data are regarded as abnormal data and the classifier provides a decision boundary between the normal data and abnormal data, although the training data do not include any abnormal data.
Compared with the state-of-the-art of anomaly detection, our method does not require any assumption about the shape (e.g. hypersphere) of the decision boundary and has fewer hyper-parameters to determine. Empirical studies on benchmark datasets verify the effectiveness and superiority of our method.
\end{abstract}

\section{Introduction}
\label{sec1}
Anomaly detection (AD) is an important research problem in many areas such computer vision, machine learning, and chemical engineering~\citep{chandola2009anomaly, ramachandra2020survey, pang2021deep, ruff2021unifying}. AD aims to identify abnormal data from normal data and is usually an unsupervised learning task because the anomaly samples are unknown in the training stage. In the past decades, numerous AD methods~\citep{scholkopf1999support, scholkopf2001estimating, breunig2000lof, liu2008isolation} have been proposed. For instance, one-class support vector machine (OCSVM)~\citep{scholkopf2001estimating} maps the data into high-dimensional feature space induced by kernels and tries to find a hyperplane giving possibly maximal distance between the normal data and the origin. \cite{tax2004support} proposed a method called support vector data description (SVDD), which finds the smallest hyper-sphere encasing the normal data in the high-dimensional feature space. SVDD is similar to OCSVM with Gaussian kernel function.

Classical anomaly detection methods such as OCSVM and SVDD are generally not suitable for large-scale data due to the high computational costs, and are not effective to deal with more complex data such as those in vision scenarios. To address these issues, a few researchers~\citet{erfani2016high, golan2018deep, abati2019latent, wang2019multivariate, qiu2021neural} attempted to take advantages of deep learning~\citep{lecun2015deep, goodfellow2016deep} to improve the performance of anomaly detection. One typical way is to use deep auto-encoder or its variants\citep{vincent2008extracting, kingma2013auto, pidhorskyi2018generative, wang2021auto} to learn effective data representation or compression models. Auto-encoder and its variants have achieved promising performance in AD. In fact, these methods do not explicitly define an objective for anomaly detection. In stead, they usually use the data reconstruction error as a metric to detect anomaly.

There are a few approaches to building deep anomaly detection objectives or models. Typical examples include classical one-class learning based approach\citep{ruff2018deep, ruff2019deep, perera2019learning, bhattacharya2021fast}, probability estimation based approach~\citep{zong2018deep, perez2019deep, su2019robust}, and adversarial learning based approach~\citep{deecke2018image, perera2019ocgan, raghuram2021general}. For example, ~\citet{ruff2018deep} proposed deep one-class classification (DSVDD), which applies deep neural network to learn an effective embedding for the normal data such that in the embedding space, the normal data can be encased by a hyper-sphere with minimum radius. The deep autoencoding Gaussian mixture model (DAGMM) proposed by ~\citet{zong2018deep} is composed of a compression network and an estimation network based on Gaussian mixture model. The anomaly scores are described as the output energy of the estimation network. ~\citet{perera2019ocgan} proposed one-class GAN (OCGAN) to learn a latent space to represent a specific class by adversarially training the auto-encoder and discriminator. The properly trained OCGAN network can well reconstruct the specific class of data, while failing to reconstruct other classes of data.
Moreover, some latest works also explore interesting perspectives. ~\citet{goyal2020drocc} proposed the method called deep robust one-class classification (DROCC). DROCC assumes that the normal samples generally lie on low-dimensional manifolds, and regards the process of finding the optimal hyper-sphere in the embedding space as an adversarial optimization problem. ~\citet{yan2021learning} claimed that anomalous domains generally exhibit different semantic patterns compared with the peripheral domains, and proposed the semantic context based anomaly detection network (SCADN) to learn the semantic context from the masked data via adversarial learning. ~\citet{chen2022deep} proposed the interpolated Gaussian descriptor (IGD) to learn more valid data descriptions from representative normal samples rather than edge samples. ~\citet{shenkar2022anomaly} utilized contrastive learning to construct the method called generic one-class classification (GOCC) for AD on tabular data. It is worth noting that classical AD methods such as OCSVM~\citep{scholkopf2001estimating} and DSVDD \citep{ruff2018deep} require specific assumptions (e.g. hypersphere) for the distribution or structure of the normal data. The GAN-based approaches ~\citet{deecke2018image, perera2019ocgan} suffer from the instability problem of min-max optimization and have high computational costs.

In this paper, we propose a novel AD method called perturbation learning based anomaly detection (PLAD). PLAD aims to learn a perturbator and a classifier from the normal training data. The perturbator uses minimum effort to perturb the normal data to abnormal data while the classifier is able to classify the normal data and perturbed data into two classes correctly.
%The learned perturbations take two forms: additive and multiplicative, which enables to capture the structure of the data better, since most difficult anomalies are usually small perturbations on the data. Consequently, we generate a set of anomalous samples with the combination of the learned perturbation of VAE and data. Then we construct a classification network to distinguish the learned anomalous samples and normal samples. The perturbator and classifier are then jointly optimized by minimizing the cross entropy of the classifier, the discrepancy between perturbation and $\bm{0}$, and the discrepancy between multiplicative perturbation and $\bm{1}$. The idea of PLAD is somehow similar to some adversarial learning based methods, which generate anomalous samples in training to benefit anomaly detection, while the optimization problem of PLAD is non-adversarial.
The main contributions of our work are summarized as follows:
\begin{itemize}[leftmargin=*]
\item We propose a novel AD method called PLAD. PLAD does not require any assumption about the shape of the decision boundary between the normal data and abnormal data. In addition, PLAD has much fewer hyper-parameters than many state-of-the-art AD methods such as~\citep{wang2019multivariate, goyal2020drocc, yan2021learning}.
\item We propose to learn perturbations directly from the normal training data. For every training data point, we learn a distribution from which any sample can lead to a perturbation such that the normal data point is flipped to an abnormal data point.
\item Besides the conventional empirical studies on one-class classification, we investigate the performance of our PLAD and its competitors in recognizing abnormal data from multi-class normal data. These results show that our PLAD has state-of-the-art performance.
\end{itemize}

\section{Proposed method}
\label{sec2}

Suppose we have a distribution $\mathcal{D}$ of $d$ dimension and any data drawn from $\mathcal{D}$ are deemed as normal data. Now we have some training data $\mathbb{X}=\{\mathbf{x}_1,\mathbf{x}_2,\ldots,\mathbf{x}_n\}$ randomly drawn from $\mathcal{D}$ and we want to learn a discriminative function $f$ from $\mathbb{X}$ such that $f(\mathbf{x})>0$ for any $\mathbf{x}\in\mathcal{D}$ and $f(\mathbf{x})<0$ for any $\mathbf{x}\notin\mathcal{D}$. This is an unsupervised learning problem and also known as anomaly detection, where any $\mathbf{x}\notin\mathcal{D}$ are deemed as abnormal data.

In contrast to classical anomaly detection methods such as one-class SVM~\citep{scholkopf2001estimating}, deep SVDD~\citep{ruff2018deep} , and DROCC~\citep{goyal2020drocc}, in this paper, we do not make any assumption about the distribution $\mathcal{D}$. We propose to learn perturbations for $\mathbf{X}$ such that the perturbed $\mathbb{X}$ (denoted by $\widetilde{\mathbb{X}}=\{\tilde{\mathbf{x}}_1,\tilde{\mathbf{x}}_2,\ldots,\tilde{\mathbf{x}}_n\}$) are abnormal but quite close to $\mathbb{X}$. To ensure the abnormality of $\widetilde{\mathbb{X}}$, we learn a classifier $f$ from  $\{\mathbb{X},\widetilde{\mathbb{X}}\}$ such that
$f(\mathbf{x})>0$ for any $\mathbf{x}\in\mathbb{X}$ and $f(\tilde{\mathbf{x}})<0$ for any $\tilde{\mathbf{x}}\in\widetilde{\mathbb{X}}$. To ensure that $\widetilde{\mathbb{X}}$ is close to $\mathbb{X}$, the perturbations should be small enough. Specifically, we propose to solve the following problem
\begin{equation}\label{obj_0}
\mathop{\text{minimize}}_{\theta,~\widetilde{\mathbb{X}}}~\frac{1}{n}\sum_{i=1}^n\ell(y_i,f_{\theta}(\mathbf{x}_i))+\frac{1}{n}\sum_{i=1}^n\ell(\tilde{y}_i,f_{\theta}(\tilde{\mathbf{x}}_i))+\frac{\lambda}{n}\sum_{i=1}^n\phi(\mathbf{x}_i,\tilde{\mathbf{x}}_i),
\end{equation}
where $\ell(\cdot,\cdot)$ denotes some loss function such as cross-entropy and $y_1=\cdots=y_n=0$ and $\tilde{y}_1=\cdots=\tilde{y}_n=1$ are the labels for the normal data and perturbed data respectively. $\lambda$ is a hyperparameter to control the magnitudes of the perturbations. $\theta$ denotes the set of parameters of the classifier $f$ and $\phi(\cdot,\cdot)$ is some distance metric quantifying the difference between two data points. However, directly solving \eqref{obj_0} encounters the following difficulties.
\begin{itemize}[leftmargin=*]
\item First, the number ($\vert\theta\vert+dn$) of decision variables to optimize can be huge if $n$ is large, where $\vert\theta\vert$ denotes the cardinality of the set $\theta$.
\item Second, it is hard to use mini-batch optimization because some decision variables (i.e.  $\widetilde{\mathbb{X}}$) are associated with the sample indices.
\item Lastly, it is not easy to determine $\phi$ because $\phi$ relies on the unknown distribution $\mathcal{D}$. For instance, $\phi(\mathbf{x},\tilde{\mathbf{x}})=\Vert \mathbf{x}-\tilde{\mathbf{x}}\Vert^2$, namely the squared Euclidean norm, does not work if $\mathbb{X}$ is enclosed by a hypersphere or hypercube (data points close to the centroid require much larger perturbations than those far away from the centroid, which implies that $\mathbf{x}-\tilde{\mathbf{x}}$ has a non-Gaussian distribution).
\end{itemize}

To overcome these three difficulties, we propose to solve the following problem instead
\begin{equation}\label{obj_1}
\begin{aligned}
\mathop{\text{minimize}}_{\theta,~\tilde{\theta}}~~&\frac{1}{n}\sum_{i=1}^n\ell(y_i,f_{\theta}(\mathbf{x}_i))+\frac{1}{n}\sum_{i=1}^n\ell(\tilde{y}_i,f_{\theta}(\tilde{\mathbf{x}}_i))+\frac{\lambda }{n}\sum_{i=1}^n\left(\Vert\boldsymbol{\alpha}_i-\mathbf{1}\Vert^2+\Vert\boldsymbol{\beta}_i-\mathbf{0}\Vert^2\right)\\
\textup{subject to}~~& \tilde{\mathbf{x}}_i=\mathbf{x}_i\odot \boldsymbol{\alpha}_i+\boldsymbol{\beta}_i, ~(\boldsymbol{\alpha}_i,\boldsymbol{\beta}_i)=g_{\tilde{\theta}}(\mathbf{x}_i), ~i=1,2,\ldots,n,
\end{aligned}
\end{equation}
where  $\mathbf{1}=[1,1,\ldots,1]^\top$ and $\mathbf{0}=[0,0,\ldots,0]^\top$ are $d$-dimensional constant vectors and $\odot$ denotes the Hadamard product of two vectors. $\boldsymbol{\alpha}_i$ and $\boldsymbol{\beta}_i$ are multiplicative and additive perturbations for $\mathbf{x}_i$ and they are generated from a perturbator $g_{\tilde{\theta}}$, where $\tilde{\theta}$ denotes the set of parameters to learn.  In \eqref{obj_1}, we hope that the multiplicative perturbation is close to 1 and the additive perturbation is close to zero but they relies on the data point $\mathbf{x}$. Therefore, the perturbation learning is adaptive to the unknown distribution $\mathcal{D}$, which solves the third aforementioned difficulty.

In fact, problem \eqref{obj_1} can be reformulated as
\begin{equation}\label{obj_2}
\begin{aligned}
\mathop{\text{minimize}}_{\theta,~\tilde{\theta}}~~&\frac{1}{n}\sum_{i=1}^n\Big(\ell\big(y_i,f_{\theta}(\mathbf{x}_i)\big)+\ell\big(\tilde{y}_i,f_{\theta}(\mathbf{x}_i\odot g_{\tilde{\theta}}^{\alpha}(\mathbf{x}_i)+g_{\tilde{\theta}}^{\beta}(\mathbf{x}_i))\big)\Big)\\
&+\frac{\lambda}{n} \sum_{i=1}^n\left(\Vert g_{\tilde{\theta}}^{\alpha}(\mathbf{x}_i)-\mathbf{1}\Vert^2+\Vert g_{\tilde{\theta}}^{\beta}(\mathbf{x}_i)-\mathbf{0}\Vert^2\right),
\end{aligned}
\end{equation}
where $\left[\begin{matrix}
g_{\tilde{\theta}}^{\alpha}(\mathbf{x}_i)\\
g_{\tilde{\theta}}^{\beta}(\mathbf{x}_i)
\end{matrix}\right]
=g_{\tilde{\theta}}(\mathbf{x}_i)$, $i=1,2,\ldots,n$. We see that we only need to optimize the parameters $\theta$ and $\tilde{\theta}$ and the total number of decision variables is $\vert\theta\vert+ \vert\tilde{\theta}\vert$, which solved the first difficulty we discussed previously. On the other hand, $\theta$ and $\tilde{\theta}$ are not associated with the sample indices, which solved the second difficulty. Once $f_{\theta}$ and $g_{\tilde{\theta}}$ are learned, we can then use $f_{\theta}$ to detect whether a new data point $\mathbf{x}_{\textup{new}}$ is normal (e.g. $f_{\theta}(\mathbf{x}_{\textup{new}})<0.5$) or abnormal (e.g. $f_{\theta}(\mathbf{x}_{\textup{new}})>0.5$). We call the method \textit{Perturbation Learning based Anomaly Detection} (PLAD).

\begin{figure}[t]
\centering
\includegraphics[width=5.5in]{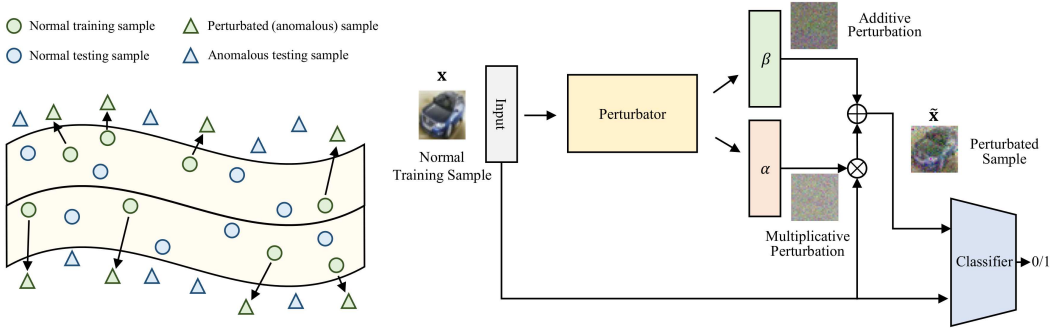}\\
\caption{The motivation and network architecture of the proposed PLAD method.}
\label{Network}
\end{figure}

In PLAD, namely (3), both the classifier $f_{\theta}$ and the perturbator $g_{\tilde{\theta}}$ are neural networks. They can be fully connected neural networks, convolutional neural networks (for image data anomaly detection), or recurrent neural networks (for sequential data anomaly detection). Figure \ref{Network} shows the motivation of our PLAD and the network architecture. Compared with many popular anomaly detection methods, our PLAD has the following characteristics.
\begin{itemize}[leftmargin=*]
\item PLAD does not make any assumption about the distribution or structure of the normal data. In contrast, one-class SVM~\citep{scholkopf2001estimating}, deep SVDD~\citep{ruff2018deep}, and DROCC~\citep{goyal2020drocc} make specific assumptions about the distribution of the normal data, which may be violated in real applications. For example, deep SVDD assumes that the normal data are encased by a hypersphere, which is hard to guarantee in real applications and may require a very deep neural network to transform a non-hypersphere structure to a hypersphere structure.
\item In PLAD, besides the network structures, we only need to determine one hyperparameter $\lambda$, which provides huge convenience in real applications. In contrast, many state-of-the-art AD methods such as~\citep{wang2019multivariate, goyal2020drocc, yan2021learning} have at least two key hyperparameters. For example, in DROCC~\citep{goyal2020drocc}, one has to determine the radius of hyper-sphere, the step size of gradient-ascent, and two regularization parameters.
\item In PLAD, we can use gradient based optimizer such as Adam to solve the optimization. The time complexity is comparable to that of vanilla deep neural networks (for classification or representation). On the contrary, many recent advances DROCC~\citep{goyal2020drocc} of anomaly detection especially those GAN based methods~\citep{deecke2018image, perera2019ocgan, yan2021learning} have much higher computational cost.
\end{itemize}

In the left of Figure \ref{Network}, actually, a normal training data can be perturbed to an abnormal data by different perturbations. Therefore, we propose to learn a distribution from which any perturbations can perturb the normal training data to be abnormal. Specifically, for every $\mathbf{x}_i$, there exists a distribution $\Omega_i$, such that for any $\mathbf{z}\in\Omega_i$, the perturbation given by $(\boldsymbol{\alpha},\boldsymbol{\beta})=h(\bm{z})$ can perturb $\mathbf{x}_i$ to be abnormal, where $h$ is a nonlinear function modelled by a neural network. We can just assume that $\Omega_i$ is a Gaussian distribution with mean $\boldsymbol{\mu}_i$ and variance $\boldsymbol{\sigma}_i^2$, i.e., $\Omega_i=\mathcal{N}(\boldsymbol{\mu}_i,\boldsymbol{\sigma}_i^2)$, because of the universal approximation ability of $h$. We take the idea of variational autoencoder (VAE) \citep{kingma2013auto} and minimize
\begin{equation}\label{eq_vae}
\mathcal{L}(\tilde{\theta}_1, \tilde{\theta}_2)=D_{K L}\left(q_{\tilde{\theta_1}}(\mathbf{z} \vert \mathbf{x}) \| p(\mathbf{z})\right)-\mathbb{E}_{q_{\tilde{\theta}_2}(\mathbf{z} \mid \mathbf{x})}\left[\log p_{\tilde{\theta}_2}(\boldsymbol{\varepsilon} \vert \mathbf{z})\right],
\end{equation}
where $p(\mathbf{z})=\mathcal{N}(\mathbf{0},\mathbf{1})$,
$\boldsymbol{\varepsilon}=\left[\begin{matrix}
\mathbf{1}\\ \mathbf{0}
\end{matrix}\right]$, $\mathbf{z}=l_{\tilde{\theta}_1}(\mathbf{x})$, $(\boldsymbol{\alpha},\boldsymbol{\beta})=h_{\tilde{\theta}_2}(\mathbf{z})$, $\tilde{\theta}_1$ are the parameters of the encoder $l$, and $\tilde{\theta}_2$ are the parameters of the decoder $h$. Combing \eqref{eq_vae} with \eqref{obj_2}, we have $\tilde{\theta}=\{\tilde{\theta}_1,\tilde{\theta}_2\}$ and solve

\begin{equation}\label{obj_3}
\begin{aligned}
\mathop{\text{minimize}}_{\theta,~\tilde{\theta}}~~&\frac{1}{n}\sum_{i=1}^n\Big(\ell\big(y_i,f_{\theta}(\mathbf{x}_i)\big)+\ell\big(\tilde{y}_i,f_{\theta}(\mathbf{x}_i\odot g_{\tilde{\theta}}^{\alpha}(\mathbf{x}_i)+g_{\tilde{\theta}}^{\beta}(\mathbf{x}_i))\big)\Big)\\
&+\frac{1}{n} D_{K L}\left(q_{\tilde{\theta_1}}(\mathbf{z} \vert \mathbf{x}) \| p(\mathbf{z})\right)-\lambda\mathbb{E}_{q_{\tilde{\theta}_2}(\mathbf{z} \mid \mathbf{x})}\left[\log p_{\tilde{\theta}_2}(\boldsymbol{\varepsilon} \vert \mathbf{z})\right],
\end{aligned}
\end{equation}
where $g_{\tilde{\theta}}=h_{\tilde{\theta}_2}\circ l_{\tilde{\theta}_1}$. The training is similar to that for VAE \citep{kingma2013auto} and will not be detailed here.

It should be pointed out that the method \eqref{obj_3} is just an extension of the method \eqref{obj_2}. In \eqref{obj_3}, we want to learn a distribution for each normal training data point $\mathbf{x}_i$ such that any perturbations generated from the distribution can perturb $\mathbf{x}_i$ to be abnormal. It is expected that \eqref{obj_3} can outperform \eqref{obj_2} in real applications. The corresponding experiments are in the supplementary material.

\section{Connection with previous works}
\label{sec3}

The well-known one-class classification methods such as OCSVM~\citep{scholkopf2001estimating}, DSVDD~\citep{ruff2018deep} and DROCC~\citep{goyal2020drocc} have specific assumptions for the embedded distribution while our PLAD does not require any assumption and is able to adaptively learn a decision boundary even if it is very complex. It is also noteworthy that the idea of DROCC in identifying anomalies is similar to ours, i.e, training a classifier instead of an auto-encoder or embedding model.

Adversarial learning based methods~\citep{malhotra2016lstm, deecke2018image, pidhorskyi2018generative, perera2019ocgan} are generally constructed with auto-encoder and generative adversarial networks (GANs)~\citep{goodfellow2014generative}, and the most widely used measure of them to detect anomalies is the reconstruction error. Compared with them, we exploit the idea of VAE when producing perturbations and the detection metric is a classifier, which should be more suitable than reconstruction error for anomaly detection. On the other hand, in these adversarial learning based AD methods, the min-max optimization leads to instabilities in detecting anomaly, while the optimization of PLAD is much easier to solve. Another interesting work SCADN~\citep{yan2021learning} tries to produce negative samples by multi-scale striped masks to train a GAN, but its anomaly score still relies on reconstruction error and the production of masks has randomness or may be hard to determine in various real scenarios. Our PLAD learns perturbations adaptively from the data itself, which is convenience and reliable.

\section{Experiment}
\label{sec4}
In this section, we evaluate the proposed method in comparison to several state-of-the-art anomaly detection methods on two image datasets and two tabular datasets. Note that all the compared methods do not utilize any pre-trained feature extractors.

\subsection{Datasets and baseline methods}
\label{sec4.1}
\paragraph{Datasets description}
\begin{itemize}[leftmargin=*]
\item \textbf{CIFAR-10:} CIFAR-10 image dataset is composed of 60,000 images in total, where 50,000 samples for training and 10,000 samples for test. It includes 10 different balanced classes.
\item \textbf{Fashion-MNIST:} Fashion MNIST contains 10 different categories of grey-scale fashion style objects. The data is split into 60,000 images for training and 10,000 images for test.
\item \textbf{Thyroid:} Thyroid is a hypothyroid disease dataset that contains 3,772 samples with 3 classes and 6 attributes. We follow the data split settings of ~\citep{zong2018deep} to preprocess the data for one-class classification task.
\item \textbf{Arrhythmia:} Arrhythmia dataset consists of 452 samples with 274 attributes. Here we also follow the data split settings of ~\citep{zong2018deep} to preprocess the data.
\end{itemize}
The detailed information of each dataset is illustrated in Table~\ref{Dataset}.
\begin{table}[!htbp]
\centering
\caption{Details of the datasets used in our experiments.}
\renewcommand{\arraystretch}{1.1}
\begin{tabular}{lccccc}
\toprule
   Dataset name     &Type  &\# Total samples   &    \# Dimension                  \\
\midrule
   CIFAR-10         &Image  & 60,000           &32$\times$32$\times$3                   \\

   Fashion-MNIST  &Image  & 70,000             &28$\times$28                   \\

Thyroid           &Tabular &3,772              &6                     \\

Arrhythmia      &Tabular &452                  &274                     \\

\bottomrule

\end{tabular}
\label{Dataset}
\end{table}

\paragraph{Baselines and state-of-the-arts.} We compare our method with the following classical baseline methods and state-of-the-art methods: OCSVM~\citep{scholkopf2001estimating}, isolation forest (IF)~\citep{liu2008isolation}, local outlier factor (LOF)~\citep{breunig2000lof}, denoising auto-encoder (DAE)\citep{vincent2008extracting}, E2E-AE and DAGMM~\citep{zong2018deep}, DCN~\citep{caron2018deep}, ADGAN~\citep{deecke2018image}, DSVDD~\citep{ruff2018deep}, OCGAN~\citep{perera2019ocgan}, TQM~\citep{wang2019multivariate}, GOAD~\citep{bergman2019classification}, DROCC~\citep{goyal2020drocc}, HRN-L2 and HRN~\citep{hu2020hrn}, SCADN~\citep{yan2021learning}, IGD (Scratch)~\citep{chen2022deep}, NeuTraL AD~\citep{qiu2021neural}, and GOCC~\citep{shenkar2022anomaly}.

\subsection{Implementation details and evaluation metrics}
\label{sec4.2}
In this section, we first describe the implementation details of the proposed PLAD method. The settings for image and tabular datasets are illustrated as follows:
\begin{itemize}[leftmargin=*]
\item \textbf{Image datasets.} For image datasets (CIFAR-10 and Fashion-MNIST), we utilize the LeNet-based CNN to construct the classifier, which is same as ~\citep{ruff2018deep} and~\citep{goyal2020drocc} to provide fair comparison. And we apply the MLP-based VAE to learn the noise for data. Since both image datasets contains 10 different classes, it can be regard as 10 independent one-class classification tasks, and each task on CIFAR-10 have 5,000 training samples (6,000 for Fashion-MNIST) and 10,000 testing samples for both of them. Consequently, the choice of optimizer (from Adam~\citep{kingma2015adam} and SGD), learning rate and hyper-parameter $\lambda$ could be varies for different classes. The suggested settings of them on each experiment in this paper refer to the supplementary material.
\item \textbf{Tabular datasets.} For tabular datasets (Thyroid and Arrhythmia), we both use the MLP-based classifier and VAE in practice, and we uniformly train them with Adam optimizer and learning rate of 0.001. Besides, $\lambda$ is set to 3 for Thyroid and 2 for Arrhythmia.
\end{itemize}
For the competitive methods in the experiment, we report their performance directly from the following paper~\citep{hu2020hrn, goyal2020drocc, yan2021learning, qiu2021neural, chen2022deep, shenkar2022anomaly} except for DROCC, which we run the official released code to obtain the results. Due to the limitation of paper length, the details of the our network settings are provided in the supplementary material.

For the selection of evaluation metrics, we follow the previous works such as ~\citep{ruff2018deep} and~\citep{zong2018deep} to use AUC (Area Under
the ROC curve) for image datasets and F1-score for tabular datasets because the anomaly detection for image and tabular datasets has different evaluation criteria. Moreover, our method does not need pre-training like ~\citep{ruff2018deep} and others did, so we uniformly train the proposed method 5 times with 100 epochs to obtain the average performance and standard deviation. Note that we run all experiments on NVIDIA RTX3080 GPU with 32GB RAM, CUDA 11.0 and cuDNN 8.0.

\subsection{Experimental results}
\label{sec4.3}

\subsubsection{Experiment on image datasets}
\label{sec4.3.1}
Table~\ref{AUC-CIFAR10} and Table~\ref{AUC-FMNIST} summarize the average AUCs performance of the one-class classification tasks on CIFAR-10 and Fashion-MNIST, where we have the following observations:
\begin{table}[h]
\centering
\caption{Average AUCs (\%) in one-class anomaly detection on CIFAR-10. Note that for the competitive methods we only report their mean performance, while we further report the standard deviation for the proposed method. * denotes we run the official released code to obtain the results, and the best two results are marked in \textbf{bold}.}
\resizebox{\textwidth}{!}{
\renewcommand{\arraystretch}{1.5}
\setlength\tabcolsep{5pt}
\begin{tabular}{l|cccccccccc}
\toprule
Normal Class & Airplane    & \makecell[c]{Auto\\mobile} & Bird & Cat & Deer & Dog & Frog & Horse & Ship & Truck \\
\midrule
OCSVM~\citep{scholkopf2001estimating}  & 61.6 & 63.8 & 50.0 & 55.9 & 66.0 & 62.4 & 74.7 & 62.6 & 74.9 & 75.9 \\
IF~\citep{liu2008isolation}           & 66.1 & 43.7 & 64.3 & 50.5 & 74.3 & 52.3 & 70.7 & 53.0 & 69.1 & 53.2 \\
DAE\citep{vincent2008extracting}     & 41.1 & 47.8 & 61.6 & 56.2 & 72.8 & 51.3 & 68.8 & 49.7 & 48.7 & 37.8 \\
DAGMM~\citep{zong2018deep}        & 41.4 & 57.1 & 53.8 & 51.2 & 52.2 & 49.3 & 64.9 & 55.3 & 51.9 & 54.2 \\
ADGAN~\citep{deecke2018image}     & 63.2 & 52.9 & 58.0 & 60.6 & 60.7 & 65.9 & 61.1 & 63.0 & 74.4 & 64.2 \\
DSVDD~\citep{ruff2018deep}        & 61.7 & 65.9 & 50.8 & 59.1 & 60.9 & 65.7 & 67.7 & 67.3 & 75.9 & 73.1 \\
OCGAN~\citep{perera2019ocgan}     & 75.7 & 53.1 & 64.0 & 62.0 & 72.3 & 62.0 & 72.3 & 57.5 & 82.0 & 55.4 \\
TQM~\citep{wang2019multivariate} & 40.7 & 53.1 & 41.7 & 58.2 & 39.2 & 62.6 & 55.1 & 63.1 & 48.6 & 58.7 \\
DROCC*~\citep{goyal2020drocc} & 79.2 & \bf74.9 & \bf68.3 & 62.3 & 70.3 & 66.1 & 68.1 & \bf71.3 & 62.3 & \bf76.6\\
HRN-L2~\citep{hu2020hrn}     & \bf80.6 & 48.2 & 64.9 & 57.4 & \bf73.3 & 61.0 & 74.1 & 55.5 & 79.9 & 71.6 \\
HRN~\citep{hu2020hrn}        & 77.3 & 69.9 & 60.6 & \bf64.4 & 71.5 & \bf67.4 & \bf77.4 & 64.9 & \bf82.5 & 77.3 \\
\midrule
PLAD     & \makecell{\bf82.5\\(0.4)} & \makecell{\bf80.8\\(0.9)} & \makecell{\bf68.8\\(1.2)} & \makecell{\bf65.2\\(1.2)}   & \makecell{\bf71.6\\(1.1)} &\makecell{\bf71.2\\(1.6)} & \makecell{\bf76.4\\(1.9)} & \makecell{\bf73.5\\(1.0)} & \makecell{\bf80.6\\(1.8)}   & \makecell{\bf80.5\\(1.3)}\\

\bottomrule
\end{tabular}}
\label{AUC-CIFAR10}
\end{table}

\begin{table}[h]
\centering
\caption{Average AUCs (\%) in one-class anomaly detection on Fashion-MNIST. Note that for the competitive methods we only report their mean performance, while we further report the standard deviation for the proposed method. * denotes we run the official released code to obtain the results, and the best two results are marked in \textbf{bold}.}
\resizebox{\textwidth}{!}{
\renewcommand{\arraystretch}{1.5}
\setlength\tabcolsep{3pt}
\begin{tabular}{l|cccccccccc}
\toprule
Normal Class & T-shirt    & Trouser & Pullover & Dress & Coat & Sandal & Shirt & Sneaker & Bag & \makecell[c]{Ankle\\boot} \\
\midrule
OCSVM~\citep{scholkopf2001estimating}        & 86.1 & 93.9 & 85.6 & 85.9 & 84.6 & 81.3 & 78.6 & 97.6 & 79.5 & 97.8 \\
IF~\citep{liu2008isolation}           & 91.0 & 97.8 & 87.2 & 93.2 & 90.5 & 93.0 & 80.2 & 98.2 & 88.7 & 95.4 \\
DAE\citep{vincent2008extracting}      & 86.7 & 97.8 & 80.8 & 91.4 & 86.5 & 92.1 & 73.8 & 97.7 & 78.2 & 96.3 \\
DAGMM~\citep{zong2018deep}        & 42.1 & 55.1 & 50.4 & 57.0 & 26.9 & 70.5 & 48.3 & 83.5 & 49.9 & 34.0 \\
ADGAN~\citep{deecke2018image}      & 89.9 & 81.9 & 87.6 & 91.2 & 86.5 & 89.6 & 74.3 & 97.2 & 89.0 & 97.1 \\
DSVDD~\citep{ruff2018deep}        & 79.1 & 94.0 & 83.0 & 82.9 & 87.0 & 80.3 & 74.9 & 94.2 & 79.1 & 93.2 \\
OCGAN~\citep{perera2019ocgan}       & 85.5 & 93.4 & 85.0 & 88.1 & 85.8 & 88.5 & 77.5 & 93.9 & 82.7 & 97.8 \\
TQM~\citep{wang2019multivariate}  & 92.2 & 95.8 & \bf89.9 & 93.0 & \bf92.2 & 89.4 & \bf84.4 & 98.0 & \bf94.5 & 98.3 \\
DROCC*~\citep{goyal2020drocc}     &88.1 &97.7 &87.6 &87.7 &87.2 &91.0 &77.1 &95.3 &82.7 &95.9 \\
HRN-L2~\citep{hu2020hrn}    & 91.5 & 97.6 & 88.2 & 92.7 & 91.0 & 71.9 & 79.4 & \bf98.9 & 90.8 & \bf98.9 \\
HRN~\citep{hu2020hrn}        & \bf92.7 & \bf98.5 & 88.5 & \bf93.1 & 92.1 & \bf91.3 & 79.8 & \bf99.0 & \bf94.6 & 98.8 \\
\midrule
PLAD     & \makecell{\bf93.1\\(0.5)} & \makecell{\bf98.6\\(0.2)} & \makecell{\bf90.2\\(0.7)}  & \makecell{\bf93.7\\(0.6)}   & \makecell{\bf92.8\\(0.8)} &\makecell{\bf96.0\\(0.4)} & \makecell{\bf82.0\\(0.6)} & \makecell{98.6\\(0.3)} & \makecell{90.9\\(1.0)}   & \makecell{\bf99.1\\(0.1)}\\
\bottomrule
\end{tabular}}
\label{AUC-FMNIST}
\end{table}

\begin{itemize}[leftmargin=*]
\item Compared with some classical shallow model-based approaches such OCSVM and IF, the proposed PLAD method significantly outperforms them on each one-class classification task with a large margin. This is mainly due to the powerful feature learning capability of deep neural network.
\item PLAD also explicitly outperforms several well-known deep anomaly detection methods such as DAGMM and DSVDD, and consistently obtains the top two AUC scores on most classes of CIFAR-10 and Fashion-MNIST compared to some latest methods such as TQM, DROCC and HRN. Specifically, in the class ``Automobile'' of CIFAR-10 and class ``Sandal'' of Fashion-MNIST, the proposed method exceeds 5.9\% and 4.7\% in terms of AUC compared to the runner-up.
\item It is noteworthy that some deep anomaly detection methods such as DSVDD and DROCC, are mainly based on the assumption that the normal data in the embedding space are situated in a hyper-sphere, while anomalies are outside the sphere. Therefore the edge of the hyper-sphere is then the decision boundary learned by the model to identify anomalies. In contrast, PLAD does not require any assumption about the shape of the decision boundary. It attempts to learn the perturbation from data itself by neural network and construct the anomalies by enforcing the perturbation to original data, then train network to distinguish the normal samples and anomalies. Of course, this is also natural to associate PLAD with some adversarial learning based methods such as ADGAN and OCGAN, etc. In contrast to them, the optimization of PLAD is a non-adversary problem, because it tries to simultaneously minimize the perturbation to produce the anomalies that similar to normal ones and minimize the cross-entropy to distinguish them.
\end{itemize}

Moreover, Table~\ref{AUC-mean} also shows the average performance on CIFAR-10 and Fashion-MNIST over all 10 classes to provide an overall comparison. Note that we further compare with two latest methods SCADN and IGD (Scratch). They do not included in the above two tables because the performance on each class is not provided in the paper. From this table we can observe that PLAD achieves the best average AUCs on both datasets among all competitive methods.

\begin{table}[h]
\centering
\caption{Average AUCs (\%) over all 10 classes on CIFAR-10 and Fashion-MNIST. Note that the best two results are marked in \textbf{bold}.}
%\resizebox{\textwidth}{!}{
\setlength\tabcolsep{7pt}
\renewcommand{\arraystretch}{1.05}
\begin{tabular}{l|cc}
\toprule
Data set  & CIFAR-10 & Fashion-MNIST \\
\midrule
OCSVM~\citep{scholkopf2001estimating}&64.7    &87.0     \\
IF~\citep{liu2008isolation}          &59.7    &91.5     \\
DAE\citep{vincent2008extracting}     &53.5    &88.1     \\
DAGMM~\citep{zong2018deep}           &53.1    &51.7     \\
ADGAN~\citep{deecke2018image}        &62.4    &88.4     \\
DSVDD~\citep{ruff2018deep}           &64.8    &84.7     \\
OCGAN~\citep{perera2019ocgan}        &65.6    &87.8     \\
TQM~\citep{wang2019multivariate}     &52.1    &92.7     \\
DROCC*~\citep{goyal2020drocc}        &69.9    &89.0         \\
HRN-L2~\citep{hu2020hrn}             &66.6    &90.0     \\
HRN~\citep{hu2020hrn}                &71.3    &\bf92.8     \\
SCADN~\citep{yan2021learning}        &66.9    &---     \\
IGD (Scratch)~\citep{chen2022deep}    &\bf74.3    &92.0     \\
\midrule
PLAD        & \bf75.1  & \bf93.5 \\
\bottomrule
\end{tabular}
\label{AUC-mean}
\end{table}

\begin{figure} \centering
\subfigure[Trouser] {
 \label{Trouser}
\includegraphics[width=0.31\columnwidth]{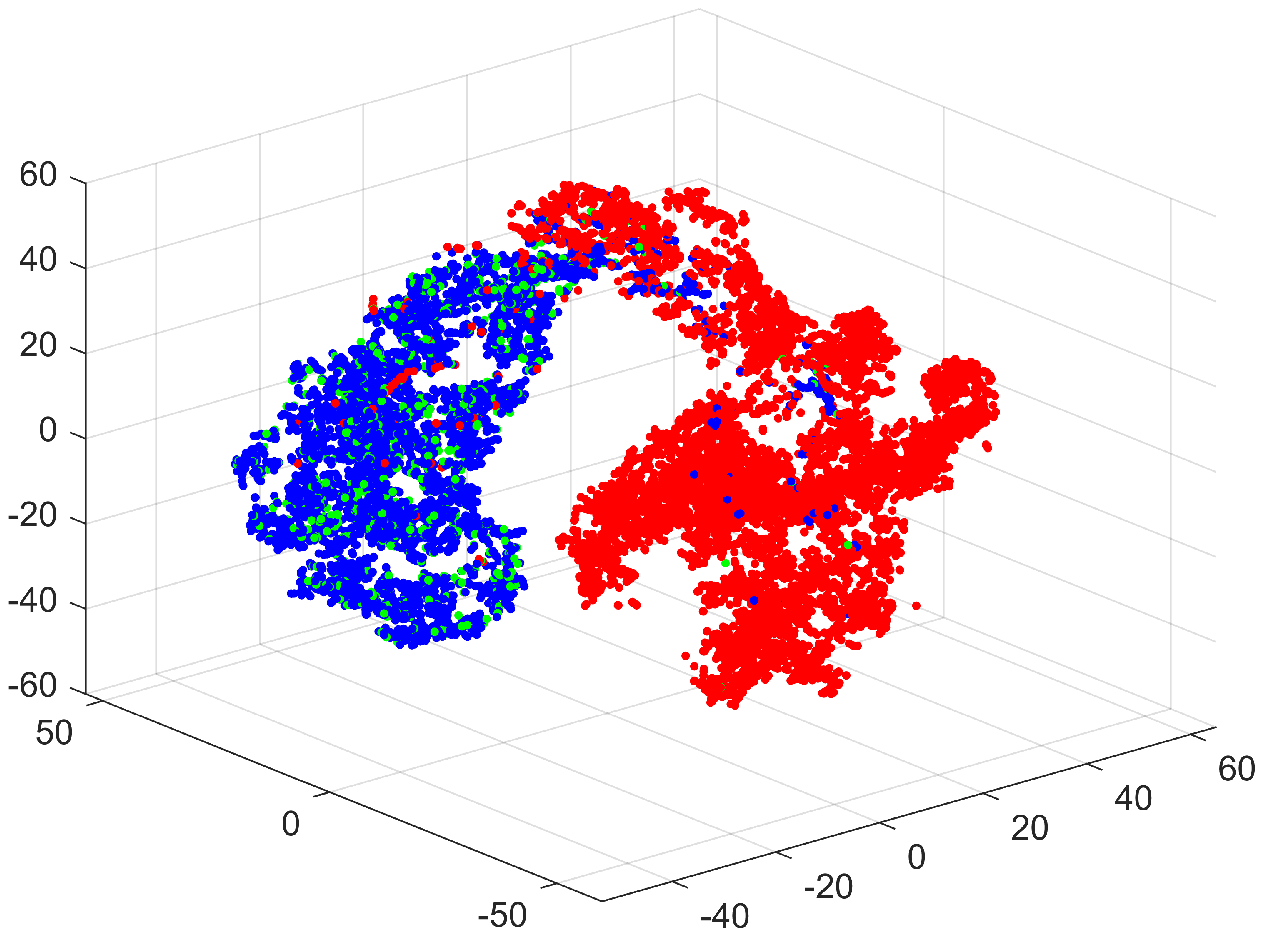}
}
\subfigure[Sneaker] {
\label{Sneaker}
\includegraphics[width=0.31\columnwidth]{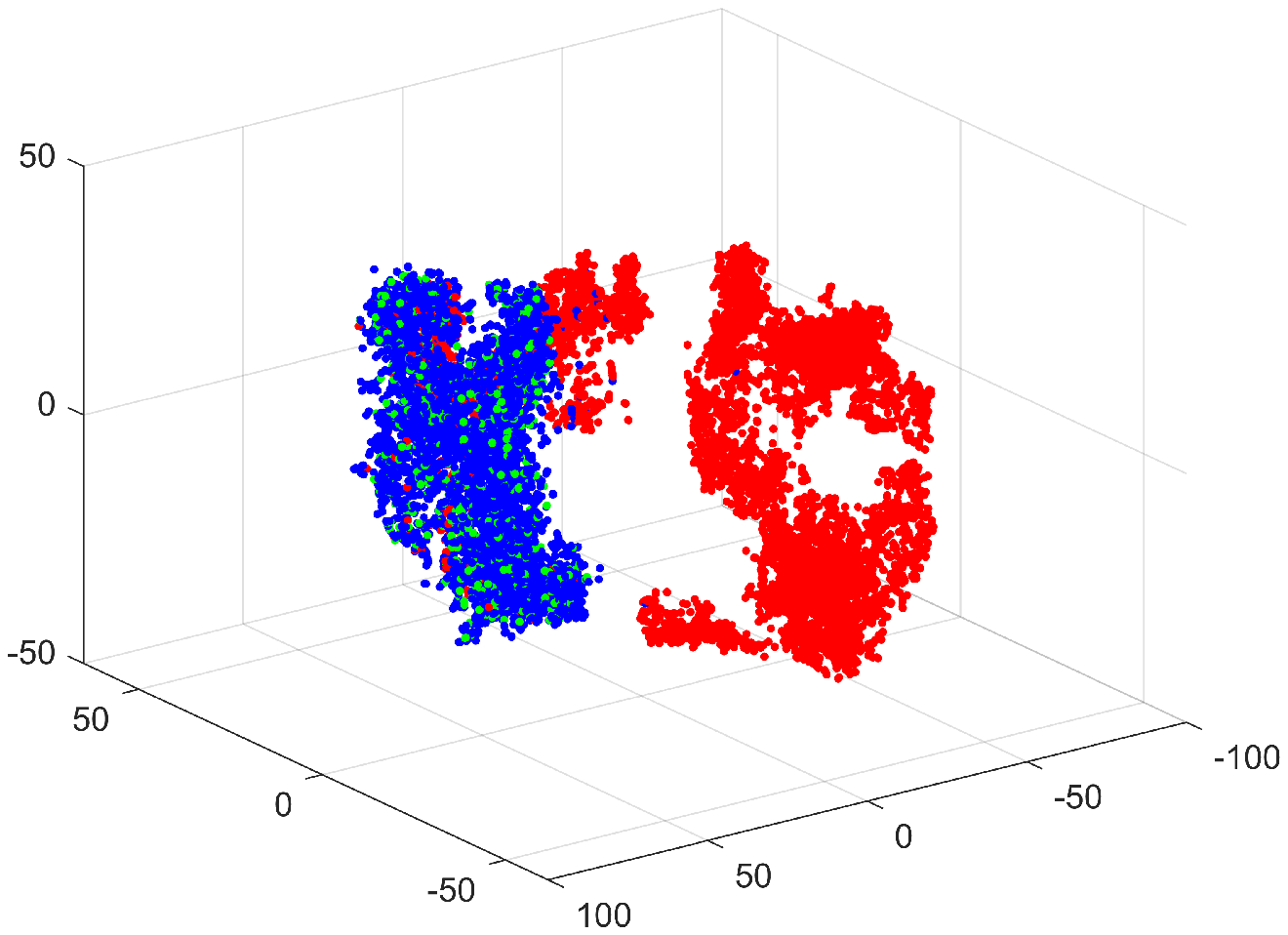}
}
\subfigure[Ankle boot] {
\label{Ankle boot}
\includegraphics[width=0.31\columnwidth]{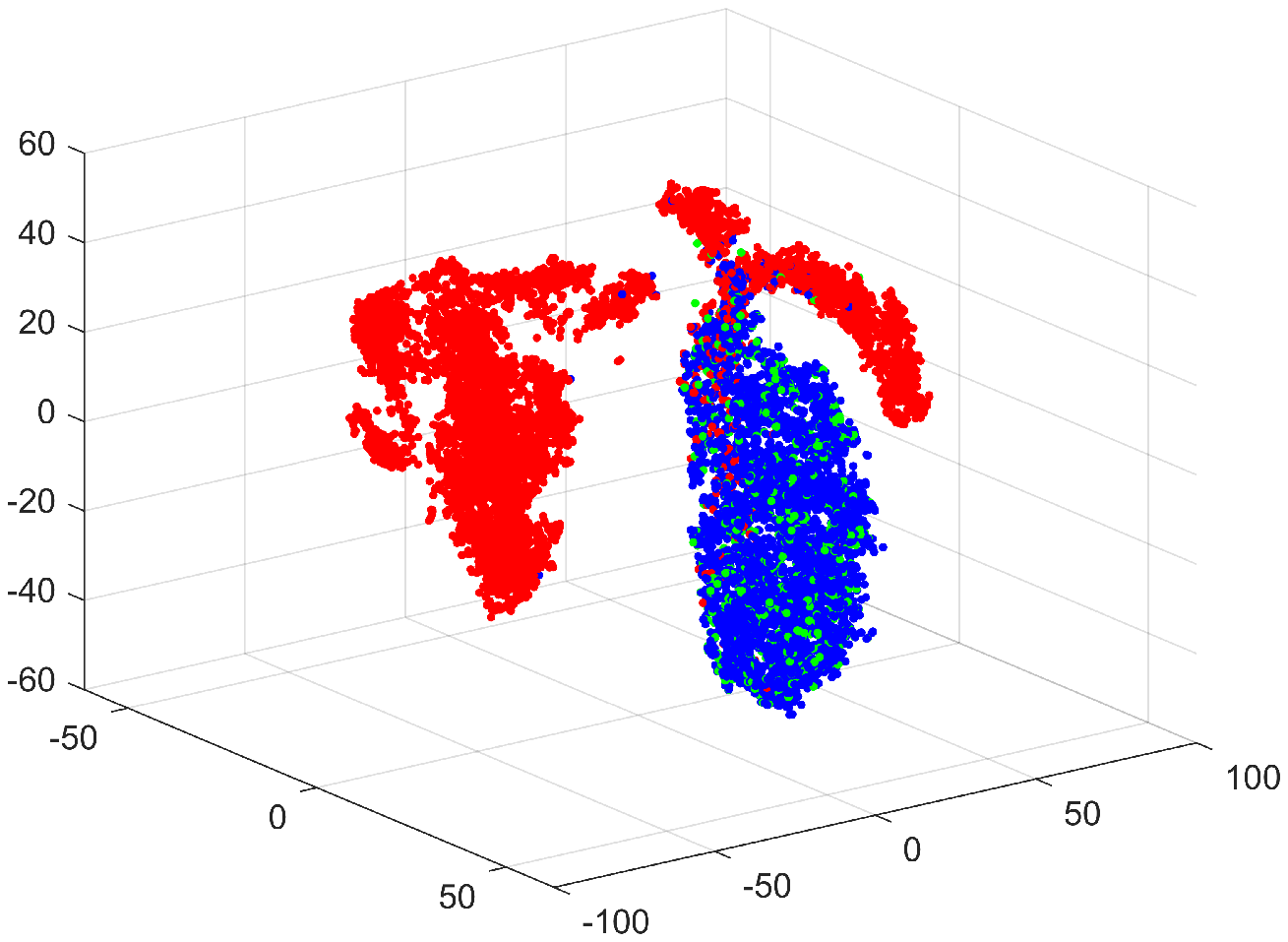}
}
\caption{The learned embedded space visualized by t-SNE. Note the points marked in  \textcolor{blue}{blue}, \textcolor{green}{green}, and \textcolor{red}{red} are training samples, normal test samples, and anomalous test samples, respectively. We select three calsses of `Trouser', `Sneaker' and `Ankle boot' in Fashion-MNIST to provide the visualization. More results for the other categories can be found on the supplementary material.}
\label{Visualization}
\end{figure}

For more intuitively understanding the learning of PLAD, we also apply t-SNE~\citep{van2008visualizing} to show the learned embedded space in Figure~\ref{Visualization}. Specifically, we use the training samples of a specific category as well as all test samples to conduct this study. Figure~\ref{Visualization} shows the visualization of the three categories `Trouser', `Sneaker' and `Ankle boot' in Fashion-MNIST. The results of the other categories can be found on the supplementary material. From this figure we can observe that the normal test samples lie in the same manifold as the training samples, while the abnormal samples are relatively separated. In other words, PLAD explicitly learns a discriminative embedded space to distinguish normal samples and anomalies.

\subsubsection{Experiment on non-image datasets}
\label{sec4.3.2}
Table~\ref{AUC-tabular} summarizes the F1-scores of each competitive method on the Thyroid and Arrhythmia datasets.
It can be observed that PLAD significantly outperforms several baseline methods such as OCSVM, DAGMM, DSVDD and DROCC with a large margin. Although NeuTraL AD and GOCC achieve encouraging 76.8\% F1-score on Thyroid, it is worth mentioning that they are both methods designed for non-image data. The Arrhythmia dataset seems to be a more difficult anomaly detection task because of its small sample size, which is not conducive to deep learning. Surprisingly, the proposed PLAD method show remarkable performance on Arrhythmia, which surpasses 9.2\% compared to the runner-up. Moreover, the performance of PLAD is also comparable to NeuTraL AD and GOCC on Thyroid, which fully demonstrates its applicability to the anomaly detection task for non-image data.

\begin{table}[h]
\centering
\caption{Average F1-scores (\%) with the standard deviation per method on two tabular datasets (Thyroid and Arrhythmia).  * denotes we run the official released code to obtain the results, and the best two results are marked in \textbf{bold}.}
%\resizebox{\textwidth}{!}{
\setlength\tabcolsep{15pt}
\renewcommand{\arraystretch}{1.2}
\begin{tabular}{l|cc}
\toprule
Data set  & Thyroid & Arrhythmia \\
\midrule
OCSVM~\citep{scholkopf2001estimating}  & 39.0 $\pm$ 1.0 & 46.0 $\pm$ 0.0  \\
LOF~\citep{breunig2000lof}        & 54.0 $\pm$ 1.0  & 51.0 $\pm$ 1.0 \\
E2E-AE\citep{zong2018deep}     & 13.0 $\pm$ 4.0 &  45.0 $\pm$ 3.0 \\
DCN~\citep{caron2018deep}      &  33.0 $\pm$ 3.0 & 38.0 $\pm$ 3.0  \\
DAGMM~\citep{zong2018deep}    &49.0 $\pm$ 4.0   &49.0 $\pm$ 3.0  \\
DSVDD~\citep{ruff2018deep}        & 73.0 $\pm$ 0.0 &  54.0 $\pm$ 1.0  \\
DROCC*~\citep{goyal2020drocc} & 68.7 $\pm$ 2.3   &   32.3 $\pm$ 1.8  \\
GOAD~\citep{bergman2019classification}  &74.5 $\pm$ 1.1   & 52.0 $\pm$ 2.3  \\
NeuTraL AD~\citep{qiu2021neural}    &\bf76.8 $\pm$ 1.9    &60.3 $\pm$ 1.1  \\
GOCC~\citep{shenkar2022anomaly}    &\bf76.8 $\pm$ 1.2    &\bf61.8 $\pm$ 1.8  \\
\midrule
PLAD        & 76.6 $\pm$ 0.6 & \bf71.0 $\pm$ 1.7 \\
\bottomrule
\end{tabular}
\label{AUC-tabular}
\end{table}

\subsubsection{Experiment on separate anomaly from multi-class normal data}
\label{sec4.3.3}
It should be pointed out that in real applications, the normal data may contains multiple classes without labels. We need to separate anomaly from these multi-class normal data. In this study, we randomly select 10,000 samples among the training split of CIFAR-10 or Fashion-MNIST to construct a new normal training set, namely, the normal data are in multiple classes. Subsequently, we randomly select two samples from the test split of CIFAR-10 or Fashion-MNIST to construct 10,000 anomalous samples using the means of pair-wise samples in pixel level. The produced anomalous samples are merged with the original test split to form a new test set. Compared with the previous two tasks (Sections~\ref{sec4.3.1} and \ref{sec4.3.2}), this one is much more difficult because the decision boundary between the anomalous samples and the normal samples are very complicated. We run the experiment to compare our method with four baselines including OCSVM, DAGMM, DSVDD and DROCC and report the results in Table~\ref{AUC-noise-extension}. We see that the proposed PLAD method outperforms the baselines significantly. Foe example, the improvement over the runner-up DSVDD is 9.1\% and 4.4\% on CIFAR-10 and Fashion-MNIST. The success of PLAD mainly stem from the ability of learning a decision boundary adaptively without any assumption.

\begin{table}[h]
\centering
\caption{Average AUCs (\%) with the standard deviation in the experiment of separating anomaly from multi-class normal data (CIFAR-10 and Fashion-MNIST).}
%\resizebox{\textwidth}{!}{
\setlength\tabcolsep{7pt}
\renewcommand{\arraystretch}{1.1}
\begin{tabular}{l|cc}
\toprule
Data set  & CIFAR-10 & Fashion-MNIST \\
\midrule
OCSVM~\citep{scholkopf2001estimating}&54.9 $\pm$ 0.0    &64.8 $\pm$ 0.0     \\
DAGMM~\citep{zong2018deep}           &44.3 $\pm$ 0.6   &49.2 $\pm$ 2.6  \\
DSVDD~\citep{ruff2018deep}           &63.6 $\pm$ 1.1    &70.9 $\pm$ 2.0     \\
DROCC~\citep{goyal2020drocc}        &60.9 $\pm$ 5.8    &68.1 $\pm$ 3.1         \\
\midrule
PLAD        & \bf72.7 $\pm$ 1.9  & \bf75.3 $\pm$ 2.8 \\
\bottomrule
\end{tabular}
\label{AUC-noise-extension}
\end{table}

\section{Conclusion}
\label{sec5}

We have presented a novel method PLAD for anomaly detection. Compared with its competitors, PLAD does not require any assumption about the distribution or structure of the normal data. This is the major reason for that PLAD outperforms its competitors. In addition, PLAD has fewer hyperparameters to determine and has lower computation cost than many strong baselines such as~\citep{wang2019multivariate, goyal2020drocc, yan2021learning}.
Actually, PLAD provides us a framework for anomaly detection. Different neural networks such as CNN~\citep{krizhevsky2012imagenet}, RNN~\citep{mikolov2010recurrent}, GNN~\citep{scarselli2008graph}, and even transformer~\cite{vaswani2017attention} can be embedded into PLAD to accomplish various anomaly detection tasks such as time series anomaly detection. One limitation of our work is that we haven't included these experiments currently.
%%%%%%%%%%%%%%%%%%%%%%%%%%%%%%%%%%%%%%%%%%%%%%%%%%%%%%%%%%%%
%%%%%%%%%%%%%%%%%%%%%%%%%%%%%%%%%%%%%%%%%%%%%%%%%%%%%%%%%%%%

\bibliographystyle{named}
\bibliography{ano}

\clearpage

\appendix

\section{Detailed settings of the network architecture, hyperparameter, and optimization}
\label{A1}
$\textbf{CIFAR-10}$\footnote{https://www.cs.toronto.edu/~kriz/cifar.html}$\textbf{:}$ We use LeNet-based classifier with 4 convolutional layers of kernel size 5 and 3 linear layers for CIFAR-10 in this paper. For the perturbator, we use fully-connected network based VAE. Note that we do not perform dimension reduction in the perturbator because our aim is to learn the perturbation from the data. We use LeaklyReLU as the activation function for both the classifier and perturbator. The detailed network structure is shown in Table~\ref{net-cifar}.

$\textbf{Fashion-MNIST}$\footnote{https://www.kaggle.com/datasets/zalando-research/fashionmnist}$\textbf{:}$ We use LeNet-based classifier with 2 convolutional layers of kernel size 5 and 3 linear layers for Fashion-MNIST in this paper. The activation function and perturbator are similar to the settings for CIFAR-10. The detailed network structure is shown in Table~\ref{net-fmnist}.

$\textbf{Thyroid}$\footnote{http://odds.cs.stonybrook.edu/thyroid-disease-dataset/} $\textbf{and Arrhythmia}$\footnote{http://odds.cs.stonybrook.edu/arrhythmia-dataset/}$\textbf{:}$  For Thyroid and Arrhythmia, we use the same fully-connected network based classifier constructed with single hidden layer. Denoting the dimensionality of input data as $d$, the size of each layer in the perturbator is then $d$. The detailed network structure is shown in Table~\ref{net-non-image}.

The one-class classification on each class can be regarded as an independent task, therefore the desirable settings of parameters for each class would be different. To improve the reproducibility of the paper, we provide some recommended settings for each parameter in Tables~\ref{settings-image} and \ref{settings-non-image}, including the hyper-parameter $\lambda$, the choice of optimizer (from Adam~\citep{kingma2015adam} and SGD), and the learning rate.

\begin{table}[!htbp]
\centering
\caption{Architecture of the LeNet-based classifier and VAE-based perturbator for CIFAR-10.}
\label{net-cifar}
\renewcommand{\arraystretch}{1.2}
\begin{tabular}{|l|}
\hline
LeNet-based Classifier                                                                                      \\ \hline
Conv2d(in\_channel=3, out\_channel=16, kernel\_size=5, bias=False, padding=2)                   \\
BatchNorm2d(16, eps=1e-4, affine=False), Leaky\_ReLU(), MaxPool2d(2,2)                           \\
Conv2d(in\_channel=16, out\_channel=32, kernel\_size=5, bias=False, padding=2)                  \\
BatchNorm2d(32, eps=1e-4, affine=False), Leaky\_ReLU(), MaxPool2d(2,2)                          \\
Conv2d(in\_channel=32, out\_channel=64, leaky\_relu\_size=5, bias=False, padding=2)                \\
BatchNorm2d(64, eps=1e-4, affine=False), Leaky\_ReLU(), MaxPool2d(2,2)                           \\
Conv2d(in\_channel=64, out\_channel=128, kernel\_size=5, bias=False, padding=2)                 \\
BatchNorm2d(128, eps=1e-4, affine=False), Leaky\_ReLU(), MaxPool2d(2,2)        \\
Flatten()                                                                                     \\
Linear(128$\times$2$\times$2, 128, bias=False)                                                                   \\
Leaky\_ReLU()                                                                   \\
Linear(128, 64, bias=False)                                                                   \\
Leaky\_ReLU()                                                                   \\
Linear(64, 1, bias=False)                                                                   \\\hline
VAE-based Perturbator              \\ \hline
Linear(3072, 3072)\\
 Leaky\_ReLU()  \\
$\bm\mu$: Linear(3072, 3072); \ \  $\bm\sigma$: Linear(3072, 3072)    \\
Reparameterzie($\bm\mu$, $\bm\sigma$) \\
Linear(3072, 3072)\\
 Leaky\_ReLU()   \\
Linear(3072, 3072$\times$2) \\\hline
\end{tabular}
\end{table}

\begin{table}[!htbp]
\centering
\caption{Architecture of the LeNet-based classifier and VAE-based perturbator for Fashion-MNIST.}
\label{net-fmnist}
\renewcommand{\arraystretch}{1.2}
\begin{tabular}{|l|}
\hline
LeNet-based Classifier                                                                                      \\ \hline
Conv2d(in\_channel=1, out\_channel=16, kernel\_size=5, bias=False, padding=2)                   \\
BatchNorm2d(16, eps=1e-4, affine=False), Leaky\_ReLU(), MaxPool2d(2,2)                           \\
Conv2d(in\_channel=16, out\_channel=32, kernel\_size=5, bias=False, padding=2)                  \\
BatchNorm2d(32, eps=1e-4, affine=False), Leaky\_ReLU(), MaxPool2d(2,2)                          \\
Flatten()                                                                                     \\
Linear(32$\times$7$\times$7, 128, bias=False)                                                                   \\
Leaky\_ReLU()                                                                   \\
Linear(128, 64, bias=False)                                                                   \\
Leaky\_ReLU()                                                                   \\
Linear(64, 1, bias=False)                                                                   \\\hline
VAE-based Perturbator              \\ \hline
Linear(784, 784) \\
Leaky\_ReLU()  \\
$\bm\mu$: Linear(784, 784); \ \  $\bm\sigma$: Linear(784, 784)\\
Reparameterzie($\bm\mu$, $\bm\sigma$) \\
Linear(784, 784) \\
Leaky\_ReLU()   \\
Linear(784, 784$\times$2) \\\hline
\end{tabular}
\end{table}

\begin{table}[!htbp]
\centering
\caption{Architecture of the MLP-based classifier and VAE-based perturbator for non-image data (Thyroid and Arrhythmia).}
\label{net-non-image}
\renewcommand{\arraystretch}{1.2}
\begin{tabular}{|l|l|}
\hline
MLP-based Classifier      & VAE-based Perturbator                                                                         \\ \hline
Input\_dim = d          &  Linear(d, d), ReLU()                                         \\
Linear(d, 20), ReLU()   &       $\bm\mu$: Linear(d, d); \ \  $\bm\sigma$: Linear(d, d)                            \\
Linear(20, 1)                                 &Reparameterzie($\bm\mu$, $\bm\sigma$) \\
                               &Linear(d, d), ReLU()   \\
                              &Linear(d, d$\times$2)\\\hline
\end{tabular}
\end{table}

\begin{table}[!htbp]
\centering
\caption{Detailed settings of the hyper-parameter $\lambda$, optimizer and learning rate for the image datasets (CIFAR-10 and Fashion-MNIST) in the one-class anomaly detection task.}
\label{settings-image}
\renewcommand{\arraystretch}{1.1}
\begin{tabular}{cccc|cccc}
\toprule
\multicolumn{4}{c|}{CIFAR-10} & \multicolumn{4}{c}{Fashion-MNIST} \\
Class      & $\lambda$  & Optimizer & Learning rate &  Class     & $\lambda$  & Optimizer & Learning rate   \\
\midrule
Airplane   & 10         & SGD       &    0.005      & T-shirt   & 5          & SGD       &    0.001    \\
Automobile & 5          & SGD       &    0.005      & Trouser   & 5          & SGD       &    0.005    \\
Bird       & 50         & SGD       &    0.005      & Pullover  & 3          & SGD       &    0.005    \\
Cat        & 5          & SGD       &    0.005      & Dress     & 3          & SGD       &    0.005    \\
Deer       & 5          & SGD       &    0.005      & Coat      & 5          & SGD       &    0.005    \\
Dog        & 10         & SGD       &    0.005      & Sandal    & 5          & SGD       &    0.005    \\
Frog       & 10         & Adam      &    0.0001     & Shirt     & 5          & SGD       &    0.005    \\
Horse      & 5          & SGD       &    0.005      & Sneaker   & 15         & SGD       &    0.002    \\
Ship       & 20         & Adam      &    0.0001     & Bag       & 5          & SGD       &    0.001    \\
Truck      & 5          & SGD       &    0.001      & Ankle boot& 5          & SGD       &    0.005    \\
\bottomrule
\end{tabular}
\end{table}

\begin{table}[!htbp]
\centering
\caption{Detailed settings of the optimizer, learning rate, and hyper-parameter $\lambda$ for the non-image datasets (Thyroid and Arrhythmia) in the one-class anomaly detection task.}
\label{settings-non-image}
\begin{tabular}{ccc|ccc}
\toprule
\multicolumn{3}{c|}{Thyroid} & \multicolumn{3}{c}{Arrhythmia} \\
$\lambda$ & Optimizer & Learning rate    &$\lambda$ & Optimizer & Learning rate    \\
\midrule
3  &Adam       & 0.001                 &2   & Adam       &    0.001      \\
\bottomrule
\end{tabular}
\end{table}

\section{Visualization of the features learned by PLAD}
\begin{figure}[h]
\centering
\includegraphics[width=5.5in]{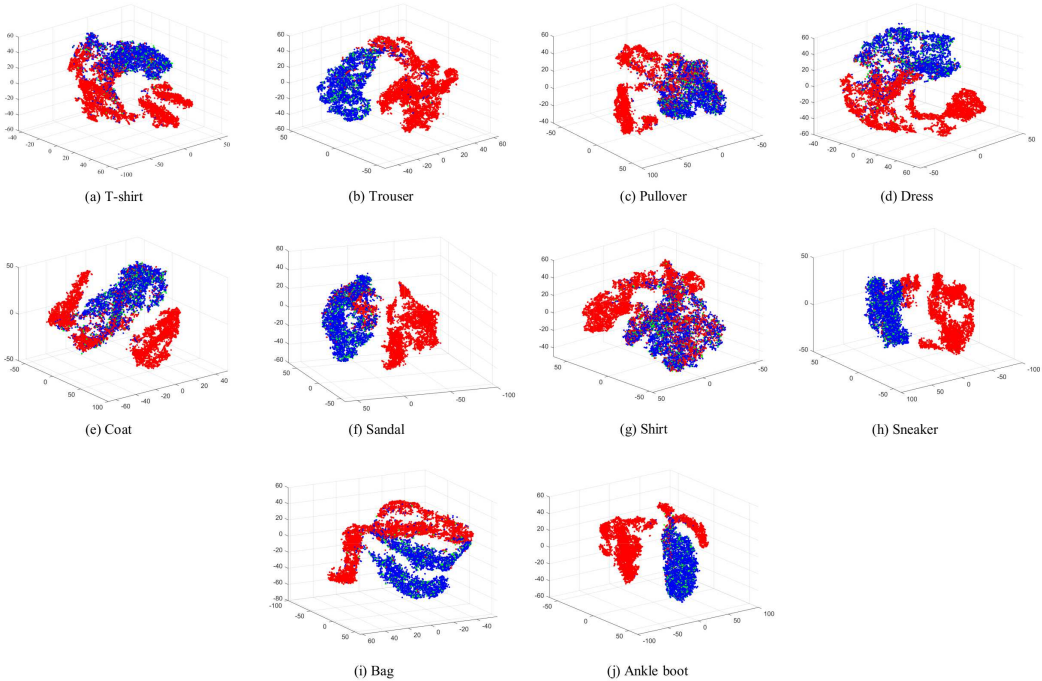}\\
\caption{Visualization of the learned embedded features of PLAD on Fashion-MNIST by t-SNE. Note the points marked in  \textcolor{blue}{blue}, \textcolor{green}{green}, and \textcolor{red}{red} are training samples, normal test samples, and anomalous test samples, respectively.}
\label{visualization}
\end{figure}

We further provide the visualization of the learned embedded features of PLAD method on each class of Fashion-MNIST in Figure~\ref{visualization}. Note that we use t-SNE~\citep{van2008visualizing} method to process the training samples of each class together with the test set, i.e., to reduce their dimensionality to 3, and they are marked in different colors. We can observe that although the learned decision boundary of PLAD does not based on any assumption, it still adaptively distinguish the normal samples and anomalies. Moreover, the training samples marked in blue and normal test samples marked in green are projected to the same space, which is consistent with our expectation of learning a space that accommodates only normal samples through the training data. Of course, the 3-D visualization shown is only for an intuitive understanding of PLAD and may not be the optimal decision boundary, so the visualization on some classes (such as Pullover and Shirt) is not desirable. PLAD may obtain a better decision boundary in higher dimension.

\section{Illustration of the anomalies produced from the multi-class normal data}
\begin{figure}[h]
\centering
\includegraphics[width=5.53in]{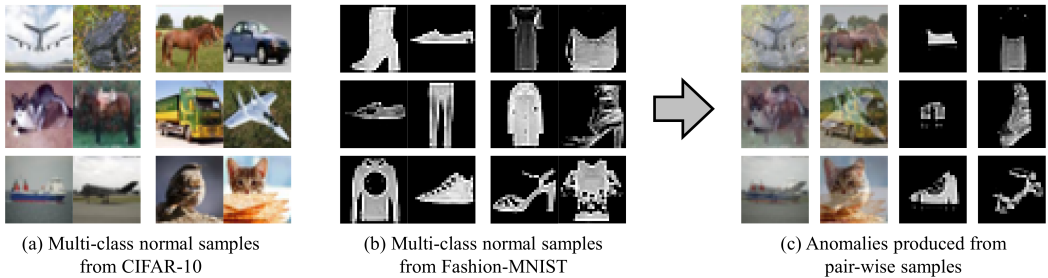}\\
\caption{Illustration of the produced anomalies from the multi-class normal data in Section~\ref{sec4.3.3}. (a) and (b) denotes the randomly selected pair-wise normal samples from CIFAR-10 and Fashion-MNIST, respectively. (c) denotes the anomalies produced by using pixel-level means of pair-wise normal samples from (a) and (b).}
\label{Multi-class-anomalies}
\end{figure}

To betted understand the experiment conducted in Section~\ref{sec4.3.3}, we show the anomalies produced from pair-wise samples in Figure~\ref{Multi-class-anomalies}. We select part of pair-wise normal samples from CIFAR-10 and Fashion-MNIST, then use their means in pixel level to produced anomalies. We can also observe from this figure that the anomalies produced in this way contain characteristics from multiple classes, which is a much more difficult anomaly detection task to solve because the decision boundary between the anomalous samples and the normal samples are very complicated.

\section{Comparison between the AE-based and VAE-based perturbators in PLAD}
We further evaluate the performance of AE-based and VAE-based perturbators. Specifically, we build an AE-based perturbator, whose architecture is same as the VAE-based one. Then we run the experiment on CIFAR-10 and Fashion-MNIST following the same experimental setup as mentioned in Section~\ref{A1}. The experimental results are shown in Table~\ref{results-ae-vs-vae}. We can observed that the VAE-based perturbator generally performs better than the AE-based one in most classes, especially in the ``Bird'' and ``Deer'' classes on CIFAR-10, and in the ``Shirt'' and ``Bag'' classes on Fashion-MNIST. Nevertheless, the AE-based perturbator still achieves remarkable performance on some classes. For example, it outperforms VAE-based perturbator on ``Truck'', ``Sandal'' and ``Ankle boot''. Overall, the average performance of AE-based and VAE-based method both outperform most competing methods in Section~\ref{sec4.3.1}
\begin{table}[h]
\centering
\caption{Comparison of the AE-based and VAE-based perturbators for the one-class classification tasks on CIFAR-10 and Fashion-MNIST.}
\label{results-ae-vs-vae}
\renewcommand{\arraystretch}{1.2}
\begin{tabular}{c|c|c|c|c|c}
\toprule
 \multicolumn{3}{c|}{CIFAR-10} &\multicolumn{3}{c}{Fashion-MNIST}\\
\midrule
Class & AE-based & VAE-based  &Class & AE-based & VAE-based\\
\midrule
Airplane   & 79.7 $\pm$ 1.2          & \bf82.5 $\pm$ 0.4         & T-shirt    & 92.3 $\pm$ 0.8         & \bf93.1 $\pm$ 0.5       \\
Automobile & 80.5 $\pm$ 0.8          & \bf80.8 $\pm$ 0.9         & Trouser    & 98.1 $\pm$ 0.5         & \bf98.6 $\pm$ 0.2        \\
Bird       & 62.3 $\pm$ 1.7          & \bf68.8 $\pm$ 1.2         & Pullover   & 88.2 $\pm$ 0.8         & \bf90.2 $\pm$ 0.7       \\
Cat        & 62.7 $\pm$ 1.5          & \bf65.2 $\pm$ 1.2         & Dress      & 92.0 $\pm$ 0.5         & \bf93.7 $\pm$ 0.6      \\
Deer       & 65.1 $\pm$ 2.8          & \bf71.6 $\pm$ 1.1         & Coat       & 91.3 $\pm$ 0.9         & \bf92.8 $\pm$ 0.8        \\
Dog        & 67.3 $\pm$ 1.5          & \bf71.2 $\pm$ 1.6         & Sandal     & \bf96.4 $\pm$ 0.3      & 96.0 $\pm$ 0.4     \\
Frog       & 72.1 $\pm$ 1.8          & \bf76.4 $\pm$ 1.9         & Shirt      & 79.1 $\pm$ 0.8         & \bf82.0 $\pm$ 0.6       \\
Horse      & 73.4 $\pm$ 1.7          & \bf73.5 $\pm$ 1.0         & Sneaker    & 98.1 $\pm$ 0.2         & \bf98.6 $\pm$ 0.3      \\
Ship       & 79.0 $\pm$ 1.4          & \bf80.6 $\pm$ 1.8         & Bag        & 86.9 $\pm$ 1.6         & \bf90.9 $\pm$ 1.0      \\
Truck      & \bf81.0 $\pm$ 0.7       & 80.5 $\pm$ 1.3            & Ankle boot & \bf99.1 $\pm$ 0.2      & 99.1 $\pm$ 0.1        \\
\midrule
Average    &72.3                     &\bf75.1                    &Average     &92.1                    &\bf93.5\\
\bottomrule
\end{tabular}
\end{table}

\section{Comparison between the CNN-based perturbator and FCN-based perturbator in PLAD}
As the above experiment can be seen, VAE-based perturbator generally performs better than AE-based one. Therefore, we further evaluate the performance of CNN-based and FCN-based VAE perturbators to provide more comprehensive analysis. Similarly, we experiment on CIFAR-10 and Fashion-MNIST. The encoder of CNN-based perturbator is similar to the classifier, with two differences that it allows bias and contains two hidden layers to produce $\bm{\mu}$ and $\bm{\sigma}$ for VAE. The dimension of hidden layer is set to 128, and the decoder is symmetric to the encoder. We show the experimental results in Table~\ref{results-vae}. We can observe that the performance of CNN-based and FCN-based perturbators are comparable in most cases, except for the more remarkable performance achieved by the FCN-based perturbator on the ``Airplane'' and ``Dog'' classes of CIFAR-10. Nevertheless, the average performance of them are quite close, and it should be noted that the CNN-based perturbator may have an encouraging performance in more complex scenarios.

\begin{table}[h]
\centering
\caption{Performance of the CNN-based and FCN-based VAE perturbators for the one-class classification tasks on CIFAR-10 and Fashion-MNIST.}
\label{results-vae}
\renewcommand{\arraystretch}{1.2}
\begin{tabular}{c|c|c|c|c|c}
\toprule
 \multicolumn{3}{c|}{CIFAR-10} &\multicolumn{3}{c}{Fashion-MNIST}\\
\midrule
Class      & CNN-based & FCN-based   &Class & CNN-based & FCN-based\\
\midrule
Airplane   & 77.6 $\pm$ 1.1          & \bf82.5 $\pm$ 0.4         & T-shirt    & 93.0 $\pm$ 0.4         & \bf93.1 $\pm$ 0.5       \\
Automobile & 80.2 $\pm$ 0.9          & \bf80.8 $\pm$ 0.9         & Trouser    & 98.4 $\pm$ 0.2         & \bf98.6 $\pm$ 0.2        \\
Bird       & 65.4 $\pm$ 1.5          & \bf68.8 $\pm$ 1.2         & Pullover   & 88.9 $\pm$ 0.3         & \bf90.2 $\pm$ 0.7       \\
Cat        & 64.6 $\pm$ 1.8          & \bf65.2 $\pm$ 1.2         & Dress      & \bf94.0 $\pm$ 0.5      &93.7 $\pm$ 0.6      \\
Deer       & \bf71.8 $\pm$ 2.3       & 71.6 $\pm$ 1.1         & Coat       & 91.9 $\pm$ 0.5         & \bf92.8 $\pm$ 0.8        \\
Dog        & 67.4 $\pm$ 1.2          & \bf71.2 $\pm$ 1.6         & Sandal     & 95.7 $\pm$ 0.4         & 96.0 $\pm$ 0.4     \\
Frog       & \bf76.6 $\pm$ 0.8       & 76.4 $\pm$ 1.9         & Shirt      & \bf82.5 $\pm$ 1.1      & 82.0 $\pm$ 0.6       \\
Horse      & 70.6 $\pm$ 1.3          & \bf73.5 $\pm$ 1.0         & Sneaker    & 98.5 $\pm$ 0.1         & \bf98.6 $\pm$ 0.3      \\
Ship       & 80.1 $\pm$ 1.7          & \bf80.6 $\pm$ 1.8         & Bag        & \bf91.0 $\pm$ 1.3      & 90.9 $\pm$ 1.0      \\
Truck      & \bf80.7 $\pm$ 1.6       & 80.5 $\pm$ 1.3            & Ankle boot & \bf99.2 $\pm$ 0.1      & 99.1 $\pm$ 0.1        \\
\midrule
Average    &73.5                     &\bf75.1                    &Average     &93.3                    &\bf93.5\\
\bottomrule
\end{tabular}
\end{table}

\section{Ablation study for PLAD}
\label{A3}
\begin{figure}[h]
\centering
\includegraphics[width=5.5in]{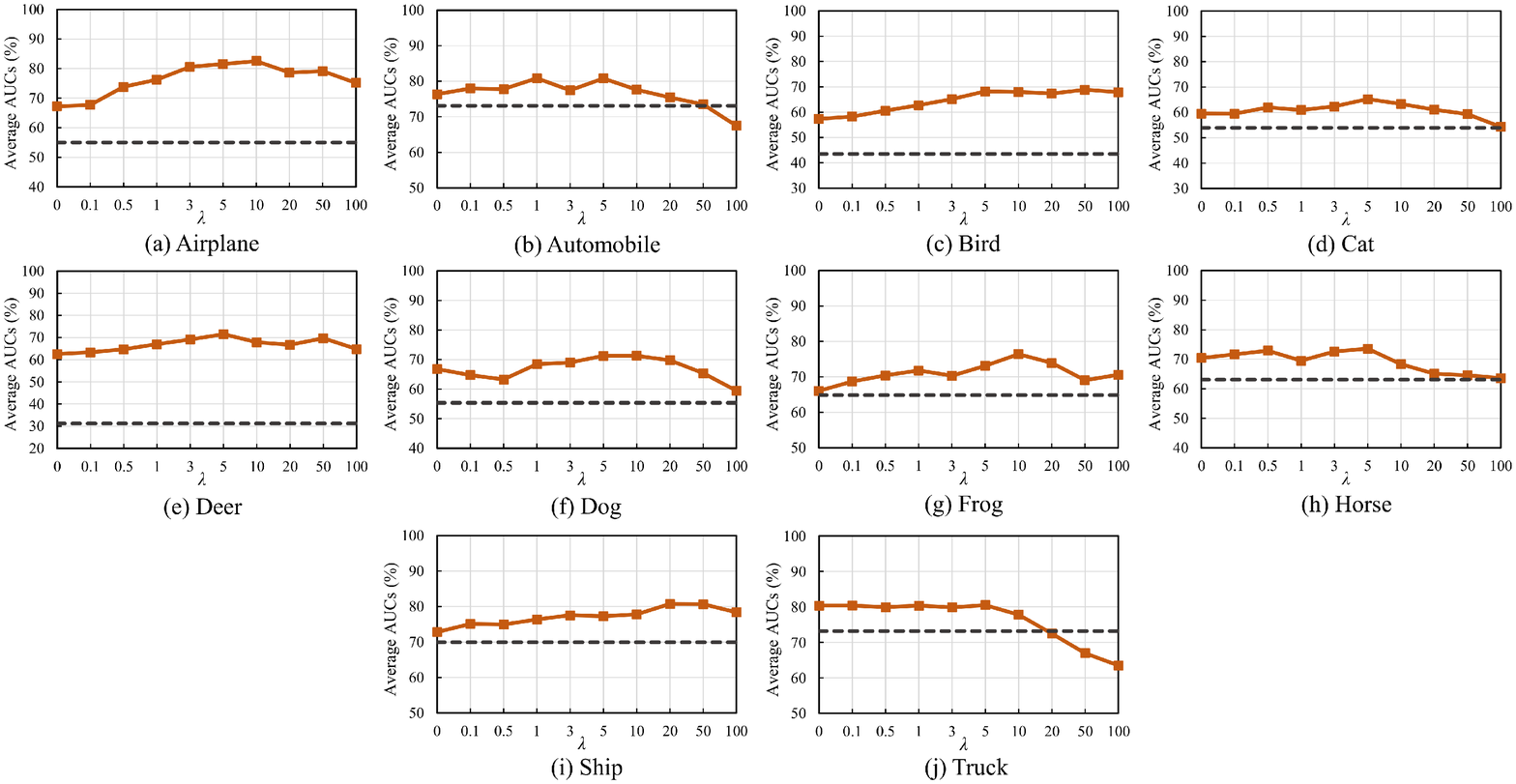}\\
\caption{Average AUCs for the one-class classification task on each class of CIFAR-10 with $\lambda$ varies in the range of $[0, 0.1, \dots, 50, 100]$. Note that the dotted line denotes the performance of the degradation model that drops the perturbator.}
\label{ablationstudy}
\end{figure}

We conduct an ablation study on CIFAR-10 to validate the effectiveness of the proposed method, and also discuss the influence of the hyper-parameter $\lambda$ to the anomaly detection performance. Specifically, we vary the value of $\lambda$ in the range of $[0, 0.1, \dots, 50, 100]$, and show the average AUCs in Figure~\ref{ablationstudy}. Note that the performance shown in dotted line indicates the degradation model of PLAD, which drops the perturber, i.e., contains only a simple classification network without any perturbation learned. We have the following observation from this figure:
\begin{itemize}[leftmargin=*]
\item The most significant AUC gap observed from Figure~\ref{ablationstudy} is wether the perturbator is considered or not. Even with very small value of $\lambda$ (e.g., $0.1$), we can see remarkable performance improvements on some classes such as ``Airplane'', ``Bird'', and `'Deer''. This indicates that the proposed perturbator forces the classifier to learn more discriminative decision boundary to distinguish the normal samples and anomalies.
\item Generally, we can observe the performance improves as the value of $\lambda$ increases. Yet, as mentioned before, anomaly detection for each class is an independent task, so the desirable range of $\lambda$ is different. For example, $\lambda \in [1, 5]$ performs well on classes ``Automobile'', ``Deer'', ``Horse'', and ``Truck'', while $\lambda \in [5, 20]$ performs well on classes ``Bird'', ``Dog'', ``Frog'', and ``Ship''.
\item We can also see that the performance commonly degrades to some extent when $\lambda$ is too large, except on the class `Frog'', where $\lambda = 100$ even performs better than $\lambda=50$. Nevertheless, the proposed PLAD method shows robustness to the variation of $\lambda$ and still achieves a relatively stable performance even against very large values.

\end{itemize}

\end{document}